\documentclass[10pt, a4paper]{article}
\usepackage[utf8]{inputenc}
\usepackage{lrec}

\usepackage{multibib}
\newcites{languageresource}{Language Resources}
\usepackage{graphicx}
\usepackage{tabularx}
\usepackage{soul}
\usepackage{enumitem}
\usepackage{xcolor}

\usepackage{tipa}
\usepackage{expex}

\usepackage{hyperref}
\usepackage{xstring}

\title{A Summary of the First Workshop on Language Technology\\ for Language Documentation and Revitalization} 

\name{Graham Neubig$^1$, Shruti Rijhwani$^1$, Alexis Palmer$^2$, Jordan MacKenzie$^3$, Hilaria Cruz$^4$, \vspace{3pt}\\
\textbf{\large Xinjian Li$^1$, Matthew Lee$^5$, Aditi Chaudhary$^1$, Luke Gessler$^3$, Steven Abney$^6$, }\vspace{3pt}\\
\textbf{\large Shirley Anugrah Hayati$^1$, Antonios Anastasopoulos$^1$, Olga Zamaraeva$^7$, Emily Prud'hommeaux$^8$,}\vspace{3pt}\\
\textbf{\large Jennette Child$^{9}$, Sara Child$^{9}$, Rebecca Knowles$^{10}$, Sarah Moeller$^{11}$,  Jeffrey Micher$^1$,}\vspace{3pt}\\
\textbf{\large Yiyuan Li$^1$, Sydney Zink$^{12}$, Mengzhou Xia$^1$, Roshan Sharma$^1$, Patrick Littell$^{10}$}\vspace{3pt}\\
$^1$Carnegie Mellon University, Pittsburgh, PA, $^2$University of North Texas, Denton, TX, \\$^3$Georgetown University, Washington, DC,$^4$University of Louisville, Louisville, KY,\\$^5$SIL International, Dallas, TX, $^6$University of Michigan, Ann Arbor, MI\\
$^7$University of Washington, Seattle, WA, $^8$Boston College, Chestnut Hill, MA,\\$^{9}$S\underline{a}nyak'ola Foundation, Port Hardy, BC, $^{10}$National Research Council Canada, Ottawa, ON, \\$^{11}$University of Colorado Boulder, Boulder, CO, 
$^{12}$Brown University, Providence, RI\\
\texttt{neubig@cs.cmu.edu, patrick.littell@nrc-cnrc.gc.ca}\\
}

\address{ }
\vspace{-12pt}

\abstract{Despite recent advances in natural language processing and other language technology, the application of such technology to language documentation and conservation has been limited. In August 2019, a workshop was held at Carnegie Mellon University in Pittsburgh to attempt to bring together language community members, documentary linguists, and technologists to discuss how to bridge this gap and create prototypes of novel and practical language revitalization technologies. This paper reports the results of this workshop, including issues discussed, and various conceived and implemented technologies for nine languages: Arapaho, Cayuga, Inuktitut, Irish Gaelic, Kidaw'ida, Kwak'wala, Ojibwe, San Juan Quiahije Chatino, and Seneca.
\\
\newline }

\begin{document}

\maketitleabstract

\section{Introduction}
Recently there have been large advances in natural language processing and language technology, leading to usable systems for speech recognition \cite{hinton2012deep,graves2013speech,hannun2014deep,amodei2016deep},
machine translation \cite{bahdanau2015neural,luong2015effective,wu2016google},
text-to-speech \cite{oord2016wavenet}, and question answering \cite{seo2017bidirectional} for a few of the world's most-spoken languages, such as English, German, and Chinese. However, there is an urgent need for similar technology for the rest of the world's languages, particularly those that are threatened or endangered. The rapid documentation and revitalization of these languages is of paramount importance, but all too often language technology plays little role in this process.

In August 2019, the first edition of a `Workshop on Language Technology for Language Documentation and Revitalization' was held at Carnegie Mellon University in Pittsburgh, PA, USA. The goal of the workshop was to take the recent and rapid advances in language technology (such as speech recognition, speech synthesis, machine translation, automatic analysis of syntax, question answering), and put them in the hands of those on the front lines of language documentation and revitalization, such as language community members or documentary linguists.

The workshop was collaborative, involving language community members, documentary and computational linguists, and computer scientists. These members formed small teams, brainstormed the future of technological support for language documentation and revitalization, and worked on creating prototypes of these technologies. Specifically, the groups focused on spoken technology (\S\ref{spoken}), dictionary extraction and management (\S\ref{dict}), supporting education with corpus search (\S\ref{corpus}), and supporting language revitalization through social media (\S\ref{social}).
%
%
These technologies were applied on nine languages, of various levels of vitality and with varying amounts of available resources:

\vspace{-6pt}
\begin{itemize}[leftmargin=4mm]\setlength{\itemsep}{-2pt}
\item \textbf{Arapaho} [arap1274], an Algonquian language spoken in the United States.
\item \textbf{Cayuga} [cayu1261], an Iroquoian language spoken in the US and Canada.
\item \textbf{Inuktitut} [inui1246], an Inuit-Yupik-Aleut language spoken in Canada. 
\item \textbf{Irish Gaelic} [iris1253], an Indo-European language spoken in Ireland.
\item \textbf{Kidaw'ida} [tait1250], a Bantu language spoken in Kenya.
\item \textbf{Kwak'wala} [kwak1269], a Wakashan language spoken in Canada.
\item \textbf{Ojibwe} [otta1242], an Algonquian language spoken in the US and Canada. 
\item \textbf{San Juan Quiahije Chatino} [sanj1283], an Otomanguean language spoken in Oaxaca, Mexico.
\item \textbf{Seneca} [sene1264], an Iroquoian language spoken in the US and Canada.
\end{itemize}

\section{Spoken Language Technology}
\label{spoken}

Most spoken language technology assumes a substantial transcribed speech corpus---on the order of hundreds or even thousands of hours of transcribed speech for a typical Automatic Speech Recognition (ASR) system \cite[for example]{hannun2014deep}.
For many languages, however, the only transcribed audio resources that exist are at the scale of minutes or an hour.

The goal of speech technology in a language revitalization setting would be to allow indigenous communities or linguists to gather corpora using up to date technological resources \cite{michaud2018integrating}. Due to the advancing age of the fluent first-language speakers in many languages, the urgency of this goal is paramount. Thus, it is important to focus on practical, labor-saving speech technologies that are feasible at the data scales that we have currently available, or at least could become feasible with further research. In this workshop, we concentrated on four speech-related tasks, each of which is feasible in low-data (and in some cases zero-data) situations. 

\subsection{First-Pass Approximate Transcription}
\label{allosaurus}

The bulk of the speech subgroup's effort concentrated on improving Allosaurus \cite{li2020allosaurus}, both improving its performance on new languages, and improving its practical usability for a real-world field context.  Allosaurus is intended as an automatic \emph{first-pass transcriber} targeting a narrow IPA representation of the audio, for the purposes of accelerating human transcription efforts or for further processing by downstream systems (e.g. approximate keyword search through an audio corpus).

Unlike a conventional ASR system, Allosaurus does not require pre-existing transcribed speech in its target language; rather, it is trained on a collection of higher-resource languages with typologically diverse phonetic inventories.  The system also differs from a conventional ASR system in that it recognizes words at the \emph{phone} level rather than the language-specific phoneme or grapheme.  

One important issue that came up was that, although the results of applying the model to a new language like Ojibwe were not unreasonable, it was difficult to imagine practical transcription in cases where the model predicted unfamiliar phones from languages with very different phonological systems. For example, many parts of a speech signal \emph{could} be identified as being voiceless vowels, which Allosaurus predicted due to having been trained on Japanese. However, presenting the user with many voiceless vowels in a language without them makes post-editing a chore.

\begin{figure}[!ht]
\centering
\includegraphics[width=\columnwidth]{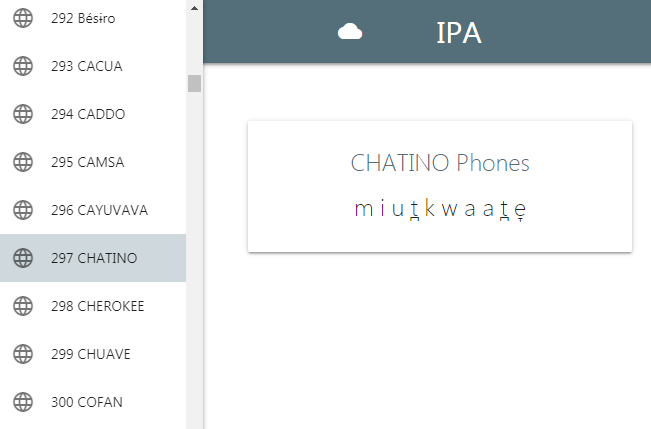} 
\caption{The user can instruct Allosaurus to restrict phone output to a particular PHOIBLE inventory.
\label{fig:allosaurus}}
\end{figure}

To mitigate this, we incorporated information from the PHOIBLE database \cite{moran2019phoible}, a large, manually-curated collection of phone inventories from roughly 2,000 languages. Given the multilingual inventory in Allosaurus (of roughly 190 sounds) and the specific inventory from some target language, as represented in PHOIBLE, we restricted the results to only include phones in the intersection of these two inventories. This both improved recognition (tests after the workshop showed 11-13\% improvement) and made human post-editing much more practical. 
In the screenshot seen in Figure \ref{fig:allosaurus}, the user has specified that they only want the recognizer to output phones in PHOIBLE's Chatino inventory.

\begin{figure}[!ht]
\centering
\includegraphics[width=\columnwidth]{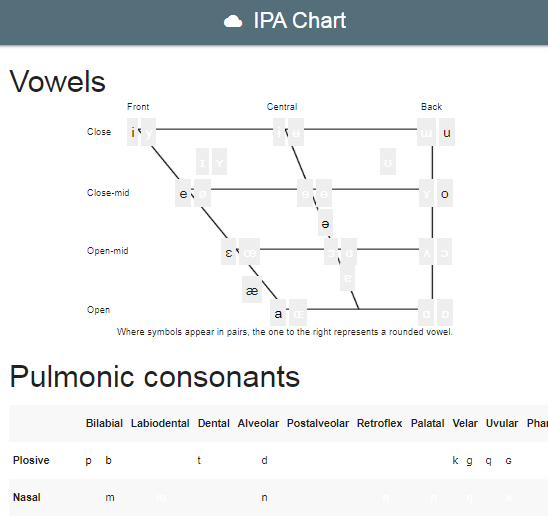} 
\caption{An interface to customize the phone inventory of a language, to restrict Allosaurus output to only those phones desired by the user.
\label{fig:allosaurus-ipa}}
\end{figure}

Additionally, we created an interface to customize the phone inventory if PHOIBLE does not provide inventory for the targeting language (Figure \ref{fig:allosaurus-ipa}). This will also allow easy testing of hypotheses for situations where the phonetic inventory is disputed.

The universal speech recognition method has already been deployed as an app that can be used for documentation at \url{https://www.dictate.app}, but quantitative testing of its  utility in an actual documentation scenario (as in \newcite{michaud2018integrating}) is yet to be performed.

Currently, Allosaurus can currently only output a phone that it has been trained to recognize, but as confused sounds frequently occur at the same place of articulation \cite{ng1998towards}, it should be possible to find links between Allosaurus's inventory and the provided inventory. Future work and possible integration with PanPhon \cite{mortensen2016panphon} could allow the tool to adapt the output to the nearest available sound (considering phonological distance) in the language's inventory.

\subsection{Phone to Orthography Decoder}

Ideally, one would like to convert the phones recognized by Allosaurus to native orthography. If successful, this would provide a speech recognition system that can directly recognize to the native orthography for low-resource languages, with minimal expertise and effort.

Many low-resource languages have fairly new orthographies that are adapted from standard scripts. Because the orthographies are new, the phonetic values of the orthographic symbols are still quite close to their conventional values. We speculate that a phonetician, given no knowledge of the language except a wordlist, could convert the phonetic representation to orthography. We made a first attempt at automating that process, in the form of an Allosaurus-to-orthography decoder called Higgins. Higgins is provided with no information about the language apart from orthographic texts and (optionally) a few pronunciation rules for cases that diverge from conventional usage of the orthography. Importantly, the texts do not include transcriptions of any recordings of interest. For now, we focus only on Latin orthography.

The pronunciation rules used to train Allosaurus were pooled and manually filtered to produce a representation of conventional pronunciation of Latin orthography. The PanPhon system \cite{mortensen2016panphon} is used to obtain a similarity metric between phonetic symbols, which we convert to a cost function for matching the phonetic symbols of the Allosaurus representation with the expected sequence for known vocabulary items. Fallback to a `spelling' mode is provided to accommodate out-of-vocabulary items. Intended, but not yet implemented, is a language model trained from the orthographic texts that the user provides. The decoding algorithm is an adapted form of the Viterbi algorithm, using beam search.

Initial results are mixed, but encouraging enough to warrant further efforts. Orthographic decoding of the output of the current acoustic model does not yield acceptably good results. However, orthographic decoding of a manual phonetic transcription of the recordings is visibly better, suggesting that our goals may be achievable with improvements in both the ASR system and Higgins itself.

\subsection{Text-to-Speech}

`Unit selection' speech synthesis is another technology that is feasible in a low-data scenario, especially when there is little or no existing transcription but a fluent speaker is available to record prompts. These systems work by identifying short segments of audio as corresponding to particular sounds, then joining these segments together in new sequences to resemble, as much as possible, fluent connected speech. Depending on the size of the target domain (i.e., what range of words/sentences the system is expected to be able to produce), intelligible speech synthesis is possible given only a few hundred or thousand utterances.

Festival \cite{blackFestival} was used to develop an initial TTS system for SJQ Chatino using a 3.8 hour corpus of Eastern Chatino of SJQ with one female speaker \citelanguageresource{cavar2016chatino} from the GORILLA language archive. Some challenges that SJQ Chatino brings for current models of speech synthesis is its complex system of tones. About 70\% of the world's languages are tonal, although tones are not represented in most orthographies. Fortunately, Chatino does represent tone on its orthography. The first output (\url{http://tts.speech.cs.cmu.edu/awb/chatino_examples/}) yielded excellent results, which is very promising. 

\subsection{Text/Speech Forced Alignment}

\begin{figure}[t]
\centering
\includegraphics[width=\columnwidth]{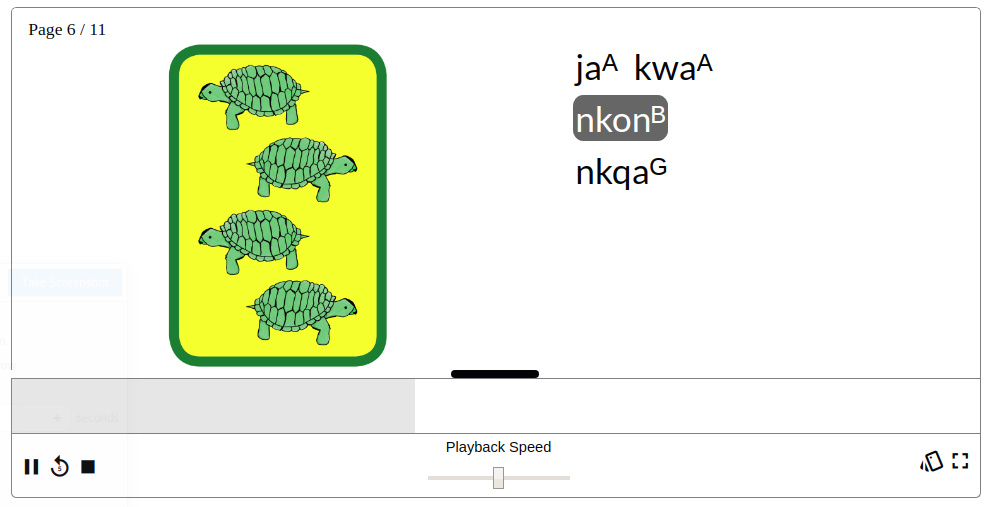} 
\vspace{-6mm}
\caption{An interactive children's book in SJQ Chatino, that highlights words when spoken in the audio, and speaks words when they are clicked.
\vspace{-3mm}
\label{fig:ras}}
\end{figure}

Text/speech forced alignment is a mature technology, and is frequently performed cross-lingually (that is, with an acoustic model that was trained on an entirely different language or set of languages), making it an obvious candidate for a technology that will work in a `no resource' situation. During the workshop, we adapted ReadalongStudio (\url{github.com/dhdaines/ReadAlong-Studio}) to align text and speech in SJQ Chatino, and produced a read-along children's book (Figure \ref{fig:ras}).

ReadalongStudio utilizes a conventional pipeline for cross-linguistic forced alignment, converting the target-language transcription so that it only uses sounds from the model language's phonetic inventory.  In this case, we used the pretrained English model included in PocketSphinx \cite{huggins2006pocketsphinx}, so the phonetic inventory in question was the English ARPABET inventory. Rather than ask the user to specify this conversion, the cross-linguistic sound-to-sound mapping is performed automatically, using PanPhon \cite{mortensen2016panphon} distances between sounds.

While the model performed adequately---a general audience would probably not notice the slight misalignment of some words---the errors revealed the importance of syllable structure to cross-linguistic alignment. When the model erred, it did so on words that had initial nasal-plosive onsets (that is, a prenasalized plosive), aligning the initial word boundary too late (that is, missing the nasal component). This is possibly because the acoustic model, trained solely on English, does not expect these sequences of sounds word-initally. For further work, therefore, we will investigate the incorporation of the multilingually-trained IPA-output model described in \S\ref{allosaurus} into our pipeline.

\section{Dictionary Management}
\label{dict}

One major issue identified in the workshop was that most existing materials (dictionaries, pedagogical grammars, etc.) are either not digitized or are not in formats that facilitate non-physical dissemination.
Even when these materials are digitized, they are typically in unstructured formats, such as unparsable PDF files, or in outdated or proprietary file formats, which are not conducive to searchability or building automatic language processing tools.

\begin{figure}[!ht]
\centering
\includegraphics[width=0.9\columnwidth]{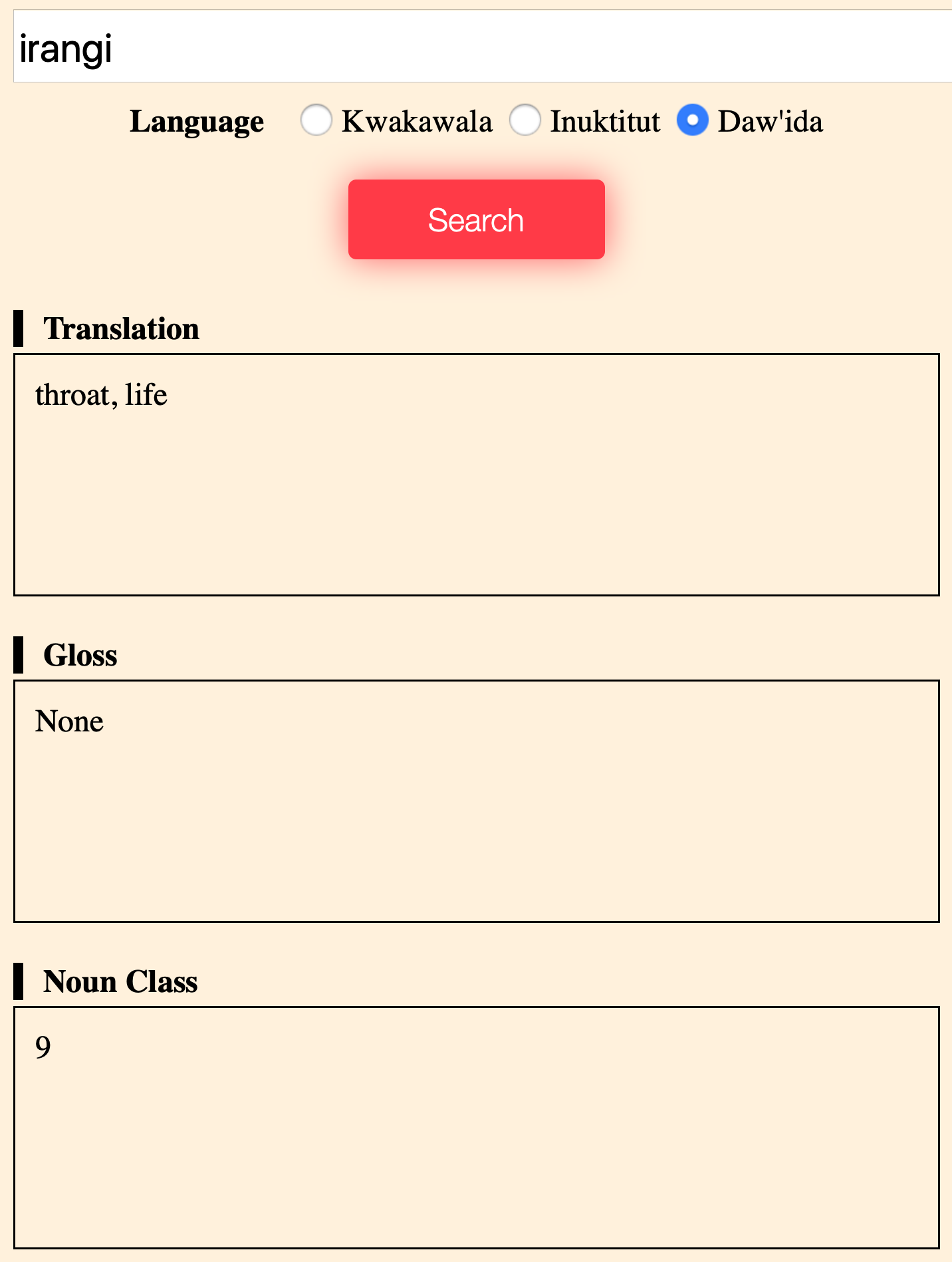} 
\caption{A schematic of the dictionary website interface.
\label{fig:dict}}
\end{figure}

Documentary linguists have long been aware of these problems and have developed best practices for data archiving and management for ongoing language documentation projects \cite{bird2003seven,himmelmann2006language,good2011data}, but the task of extracting structure from older resources remains a challenge. During the workshop, we focused on three major avenues:

\begin{itemize}\setlength{\itemsep}{0pt}
\item Converting dictionary materials from unstructured to structured formats, so that they can be easily used for natural language processing and other computational research, as well as easily presentable in interactive formats online.
\item Enhancing the annotations of such dictionaries so that they are tailored to the needs of the language community and to the intricacies of the language.
\item Creating a demonstrative website that showcases the created dictionaries along with search functionalities, to retrieve the structured information and annotations extracted from the original PDF files (see Figure \ref{fig:dict}). 
\end{itemize}

The contents of the original files, in Microsoft Word and PDF format, are not directly accessible by several existing NLP tools. We first converted these into a structured format easily retrieved by our web search service. As a first step, the files are converted to plain text documents. Then, we used regular expressions within Python scripts to extract the relevant information and store them as structured tables.

The website currently supports three languages: Kwak'wala, Inuktitut and Kidaw'ida. Users are able to search for a word in each of these languages, and the website displays its English translation, gloss, noun class, part of speech tag, parallel example sentences, words with the same root and words with the same suffix.  

For future work, we plan to expand the annotations of our dictionaries to include morphological information, potentially using automatic segmentation and morphological analysis tools. In addition, as we digitize more materials, we will expand vocabulary coverage and add sentence examples for words in a variety of contexts. Ideally, we will also be able to add audio recordings of example phrases, which is especially important for the growing number of second language learners.

\subsection{Kwak'wala}

Kwak'wala (previously known as Kwakiutl) is a Wakashan language spoken in Western Canada by the indigenous Kwakwaka'wakw people (which means ``those who speak Kwak'wala").
It is considered an endangered language since there are fewer than 200 fluent Kwak'wala speakers today (about 3\% of the Kwakwaka'wakw population).

For Kwak'wala, there exist different sources of educational materials in different formats. To make things worse, there are no versions of these materials in easily-searchable digital formats, and as a result the teachers of the language find it difficult to provide language learners with a comprehensive source of information. We processed:

\begin{itemize}[topsep=0pt,noitemsep]
\item A dictionary with the original word and its translation
\item A file with example sentences for some words
\item A file with suffix information for every word 
\end{itemize}

All of the files were in PDF format -- these were converted into computer-readable format as described above, and aggregated to be served on our website.

While some dictionaries have morphological information available that will allow for matching queries to stems or suffixes, not all do. One approach to producing automatic alternatives is to use morphological segmentation tools, including unsupervised tools that try to infer morphological boundaries from data. For Kwak'wala, we experimented with two approaches: Morfessor, a probabilistic model for doing unsupervised or semi-supervised morphological segmentation \cite{virpioja2013morfessor} and byte pair encoding, a compression model that can be applied to automatically segment words \cite{sennrich2016neural}. 
We use these models to split words up into approximate roots and affixes, which are then used for search or to cluster similar words to show to users. Note that there is scope for improvement on this front, including having language speakers determine which automatic systems produce the best segmentations, exploring different ways of visualizing related words, and using these to improve dictionary search.

\subsection{Inuktitut}

Inuktitut is a polysynthetic language spoken in Northern Canada and is an official language of the Territory of Nunavut. Inuktitut words contain many morphemes, which are realized as different surface forms depending on their context.  
As such, it is useful for a dictionary to connect different surface forms that share an underlying form, to allow morpheme glossing and comparison with morphemes in other dictionary entries. To do this, we built a prototype system using a morphological analyzer for Inuktitut called Uqailaut \cite{farley2012uqailaut}, which produces a morphological segmentation for an input token along with each morpheme's underlying form. Using dictionary files provided by the analyzer tool, we provide a gloss lookup for each morpheme.
Making use of a previously analyzed corpus of parallel Inuktitut-English legislative proceedings \cite{micher2018provenance}, we display any parallel sentences containing the input word. Finally, from a database of pre-analyzed words, the website locates additional words containing the same deep form morphemes and displays them to the user. 


\subsection{Kidaw'ida}

Kidaw'ida [dav; E.74] (also Dabida, Taita) is a Bantu language of southeast Kenya. Despite having a robust population of speakers, estimated at 275,000 as of 2009 \cite{ethnologue2019}, Kidaw'ida is highly under-documented. Furthermore, per the hegemony of Kiswahili and English in the region (even in rural areas), extensive asymmetrical borrowings from these languages have occurred, entailing some degree of language shift \cite{lugano2019kidawida}. The diachronic loss of Kidaw'ida is complicated by the fact no extensive grammatical descriptions have been written of the language, though sketches do exist \cite{wray1894elementary,sakamoto2003introduction}. Kidaw'ida lacks a comprehensive mono- or bilingual dictionary. Addressing this shortage of materials is a chief and pressing aim of any documentation effort of Kidaw'ida, with the goal of linguistic preservation and bolstering the status of the language in Kenya.

Kidaw'ida, like any Bantu language, presents a variety of challenges for lexicography because of its complex noun class system---more appropriately deemed a \emph{gender} system \cite{corbett1991}---in which extensive agreement for noun class is realized in prefixes on nouns, verbs, adjectives, prepositions, and determiners \cite{mdee1997nadharia}. Kidaw'ida has nine genders (following \newcite{carstens2008dp}), the first six of which contain singular-plural pairs. This means that the language has up to fifteen surface forms of words showing noun class agreement. For instance, the genitive `of' attests these surface forms: \textit{wa}, \textit{w'a}, \textit{ghwa}, \textit{ya}, \textit{cha}, \textit{va}, \textit{lwa}, \textit{ra}, \textit{jha}, \textit{gha}, \textit{ya}, \textit{ra}, \textit{ghwa}, \textit{kwa}, and \textit{kwa}. This example highlights a fundamental tension of Bantu lexicography: (redundant) specificity vs. (unclear) simplicity. We are faced with the choice between fifteen distinct and not clearly related entries for genitive `of' vs. one underspecified entry `-\textit{a}' that relies on the dictionary user's understanding of the morphological formalism of the hyphen denoting a bound root \emph{and} the noun class agreement system of the language. 
	
A main goal in the documentation of Kidaw'ida is to bring current materials in the language ``up to date" so they may be processed into dictionaries, grammars, and texts that can be used for pedagogical ends to preserve the language and boost the rate of formal literacy. Our main aim in the context of the workshop was converting extensive yet only semi-organized word lists into a coherent and detailed spreadsheet that can be fed into a lexicography program, such as TLex\footnote{\url{https://tshwanedje.com/tshwanelex/}}, which is well suited for Bantu languages. We focused on automating dictionary extraction processing from a word list of Kidaw'ida and English, with definitions, drafted by a native speaker of Kidaw'ida. The original files were in Microsoft Word format and converted using regular expressions as described earlier in this section.

Additionally, we attempted to computationally augment the dictionary with supplementary linguistic information: specifically, part-of-speech tags and noun classes. We used spaCy\footnote{\url{https://spacy.io}}, a natural language processing library, to automatically obtain the part-of-speech tags for the English translation for each dictionary item. These are projected onto the corresponding Kidaw'ida word or phrase. Furthermore, noun classes in Kidaw'ida can be estimated, with reasonable accuracy, by examining the prefix of words that are tagged as nouns. Using a predetermined set of prefixes, we add the predicted noun classes to each word in the dictionary. Although allomorphy is present and noun class marking can occasionally be slightly irregular, in the vast majority of cases this approach greatly reduces annotation effort.

The Kidaw'ida dictionary search on our website retrieves the predicted linguistic information in addition to the English translation of the input word. Future work might focus on the creation of platforms to facilitate user-generated data for dictionaries, grammars, and texts, in which users could record pronunciations of words and phrases. This project would also be well served by forum-type interfaces for the generation of meta-linguistic and contextual information furnished by native speakers. 

\subsection{Chatino}

The digitized resource was a collection of complete verb inflection tables in SJQ Chatino.
SJQ Chatino verb inflection depends on a rich system of tones \cite{cruz2019}. At the core of this system is an inventory of tone triplets which serve to express person/number distinctions and aspectual/modal distinctions. Each triplet contains a tone expressing first-person singular agreement, another expressing second-person singular agreement, and a third expressing other person-number agreement categories; in addition, each triplet expresses one or another category of aspect/mood. Many triplets are polyfunctional, in the sense that the aspect/mood that they expressed varies according to a verb's inflection class. 

The morphological tags were converted to match the UniMorph standard \cite{sylakglassman2016}, with annotations for number, person, and aspect. The result includes a total of 4716 inflection paradigms for more than 200 lemmata, which we used for training automatic morphological inflection systems with promising results.

The development of an SJQ orthography began in 2003 and since then, the system has undergone several changes, especially for tone notation. The current literature employs three different systems for representing tone distinctions: the S-H-M-L system of E.~\newcite{cruz2011phonology}; the numeral system of H.~\newcite{cruz2014linguistic}; and the alphabetic system of E.~\newcite{cruz2013tonal}. During the workshop we consolidated two word documents containing 210 paradigm conjugations. Each document had a different tone representation. 

\section{Teacher-in-the-Loop: Search Tools for Low-Resource Corpora}
\label{corpus}

The goal of this project is to provide an easy-to-use search interface to language corpora, with the specific aim of helping language teachers and other users develop pedagogical materials. In a recent survey of 245 language revitalization projects \cite{perez2019global}, more than 25\% of respondents selected Language Teaching as the most important objective of revitalization efforts. Language Teaching was one of 10 options (including an Other option), and more respondents selected Language Teaching as their top objective than any other option. At the same time, it is not at all clear that the typical products of endangered language documentation projects are inherently useful for language teachers. Much has been written on this issue -- here we mention only a handful of the relevant literature. \newcite{yamada2011integrating} takes a positive view of the situation, presenting a successful case study and arguing for new language documentation methods to directly support language teaching. Others \cite[for example]{miyashita2013collaborative} describe the difficulty of transforming documentation products into language teaching materials. \newcite{taylor2019recording} writes that, even when text collections or grammars are available for a language, searching them for materials suitable to a particular lesson can be a daunting task. 

Naturalistic language data has been shown to be more effective in language pedagogy than artificially constructed data, since naturalistic examples demonstrate not only how to form a word or phrase correctly but also how the phenomenon is used in real contexts, whether monologue, narrative, or conservation \cite{reppen2010using}. However, finding naturalistic data can be difficult for rare grammatical phenomena, and while major languages have mature corpus search engines like the Corpus of Contemporary American English (COCA) \citelanguageresource{davies2008corpus} or the Russian National Corpus \cite{apresjan2006russian}, low-resource languages typically lack not only graphical search interfaces, but also the rich annotations (such as morphological and syntactic parses) that are conventionally required to support the function of a search interface. Revitalization programs only rarely have the resources available to enrich a corpus with such expensive annotations, to say nothing of the technical challenge of creating a search interface once the annotation has been completed.

During the workshop, we imagined a corpus search system that would be good enough to let a language teacher find instances of a target grammatical phenomenon without requiring the corpus to be large or richly annotated, and without requiring the user to express the query in technical terms. For example, a teacher may be planning a lesson on transitive verbs and would enter as a query a sentence like \emph{Then my brother carried me down the hill}. The envisioned system, which we call Teacher-in-the-Loop (\url{https://github.com/lgessler/titl}), allows users to enter a sentence or phrase and uses the input to query an unannotated corpus for syntactically or semantically similar sentences. The system presents several examples to the user, who marks each new result as either relevant or not relevant before completing the loop by submitting their query again. The system returns more sentences, guided by the user's feedback, and the process continues. When the user has amassed enough material for their purposes, the selected sentences can be exported to a file. 

This system can be thought of as having two main components: a search engine (\S\ref{search-engine}), and a graphical user interface (\S\ref{corpus-gui}) that engages users in the feedback cycle (human-in-the-loop). The corpus search will be evaluated as successful if it can retrieve useful examples in any language. The interface must allow users to select candidates as relevant or not, use that feedback to improve the search, and export the best results in an easily accessible format. 

During the workshop, we tested our system using a corpus of Arapaho texts \citelanguageresource{cowellArapaho}. Arapaho is a severely endangered Plains Algonquian language, spoken in the western United States of Wyoming and Oklahoma. A finite state transducer model for morphology and an online lexicon and interactive online dictionary exist for Arapaho \cite{Kazeminejad2017CreatingLR}, but we do not exploit these resources for the sake of our experiment that is aimed at languages without such resources. 

\subsection{Corpus Search Engine}
\label{search-engine}

For the corpus search engine, we implemented two different methods: fuzzy string matching, which returns orthographically similar strings, and word embeddings, which are learned numerical representations of words or sentences designed to find semantically, and possibly syntactically, similar tokens/sentences. The two methods have potentially complementary strengths, and they also have dramatically different needs for data and computation. 

\textbf{Fuzzy string matching} is a pattern-matching approach for retrieving similar strings. Since it does not involve any training of machine learning models, which can often require a large amount of data, fuzzy matching can be easily applied for low-resource languages. It is appropriate when users are looking for orthographically similar, but not necessarily exactly matched strings. 

\textbf{Word embeddings} have also been successful in approximate search, 
finding semantic similarities between words even across languages. A word embedding 
is typically a vector representation, trained on large amounts (at least 1M tokens) of monolingual text, whose values reflect syntactic and semantic characteristics of the word, based on the contexts in which the word appears. These embeddings can be obtained using different algorithms, such as GloVe \cite{pennington2014glove}, fastText \cite{bojanowski2017enriching}, or BERT \cite{devlin2019bert}, among many others. 

\subsection{Graphical User Interface (GUI)}
\label{corpus-gui}

A graphical user interface is essential to the system because it makes it easy for our target users (teachers) to query the corpus. Our GUI's design is simple and language-independent. 
From the user's perspective, the GUI workflow consists of the following steps:

\begin{enumerate}\setlength{\itemsep}{0pt}
\item An initial query sentence is entered into the interface
\item The system returns a small number of relevant sentences from the corpus
\item The user marks each of the new sentences as either relevant or not relevant
\item The user asks the system for more sentences
\item Steps 2--4 are repeated until the user is satisfied, at which point they may download a file with their relevant examples
\end{enumerate}

\begin{figure}[t]
\centering
\includegraphics[scale=0.5]{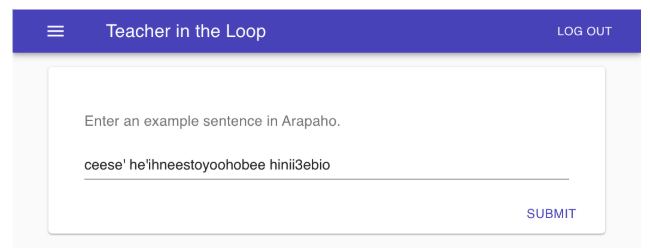} 
\vspace{-5mm}
\caption{An example of a user entering a sentence into the teacher-in-the-loop system.
\label{fig:titl1}}
\vspace{-3mm}
\end{figure}

The workflow is demonstrated 
above, with Arapaho as example language. In Fig.~\ref{fig:titl1}, we see the user entering the first query sentence. In this example, the user wants to find sentences  similar to \emph{ceese' he'ihneestoyoohobee hinii3ebio}, which means `one remained behind'. The user's intent is to find other sentences with the narrative past affix \emph{he'ih}. 

\begin{figure}[t]
\centering
\includegraphics[scale=0.45]{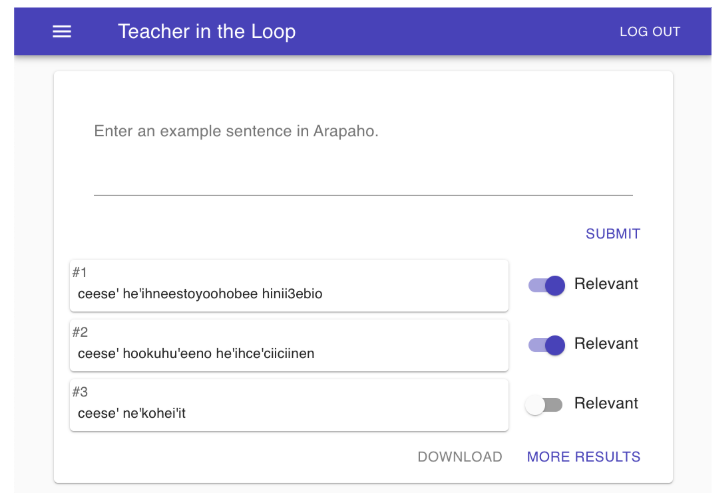} 
\vspace{-5mm}
\caption{Results from the teacher-in-the-loop system for an example sentence.
\label{fig:titl2}}
\vspace{-3mm}
\end{figure}

In Fig.~\ref{fig:titl2}, we see the state of the system after a user has queried the system for more sentences and has marked the sentences which are relevant. Sentence \#1 repeats the query sentence. Sentence \#2, \emph{ceese' hookuhu'eeno he'ihce'ciiciinen}, which means `the skulls were put down again', is marked as relevant by the user since it includes the narrative past affix \emph{he'ih}. Sentence \#3
is marked not relevant because, while it does have \emph{ceese'} in common with the query sentence, the \emph{he'ih} affix is not present. Note that this is tied to the query: if the user had instead been looking for sentences with \emph{ceese'} in it, all three sentences would have been relevant.

The user repeats this process, iteratively retrieving more sentences and marking them relevant or irrelevant, until they decide they have enough sentences, at which point they can download all of their sentences.


From the system's perspective, this interaction requires the following steps:

\begin{enumerate}\setlength{\itemsep}{0pt}
\item We compute an embedding for each sentence in the entire corpus and store it along with the sentence's text.
\item On receiving the user's first query, we convert the query into a vector. The top $k$ sentences that have the highest cosine similarity with the vector are returned. 
\item On receipt of the user's relevance judgments on the $k$ sentences that were just presented, the system:
\begin{itemize}\setlength{\itemsep}{0pt}
    \item discards the irrelevant sentences;
\item computes a new query vector by taking the mean of the vectors for sentences marked as relevant;
\item returns the next k sentences with the highest cosine similarity to the new query vector. 
\end{itemize}
\end{enumerate}

The prototype we developed at the workshop used fastText \cite{bojanowski2017enriching} to compute word vectors from an Arapaho corpus of 100,000 tokens. Sentence vectors were derived by taking the arithmetic mean of all word vectors in the sentence. Sentence vectors were stored alongside their text in files, and an HTTP service was created that can take a query and produce the next $k$ nearest sentences. 
The user interface was developed in JavaScript using the Meteor framework (\url{www.meteor.com}) and the UI libraries React (\url{reactjs.org}) and Material-UI (\url{material-ui.com}). 

Embedding algorithms typically assume that a lot of training data is available, and getting good embeddings with a small corpus has been a challenge. Another challenge is that embeddings tend to capture primarily semantic information with some syntactic information, while we want the reverse. Our team has continued to investigate variations on algorithms that might produce the best results. Approaches under consideration include transfer learning with BERT \cite{devlin2019bert}, as well as training using skip-grams over part of speech tags or dependency parses \cite{mikolov2013distributed,levy2014dependency}.


We plan to continue developing embedding strategies that are performant and syntactically rich even when trained with little data, to incorporate fuzzy string matching (possibly augmented with regular expression capabilities) into our system, and to conduct human evaluations that will assess the system's success as a search interface. 

\section{Social Media}
\label{social}

Recently, it has become prevalent for speakers or learners of endangered languages to interact with language on social media \cite{huaman2011indigenous,jones2012social}. Previous works on developing extensions for social media include \newcite{scannell2012translating}, who proposed unofficial translation for endangered languages on Facebook, and \newcite{arobba2010community} who developed a website called LiveAndTell for sharing, teaching, and learning by, of, and for Lakota speakers and communities interested in preserving Lakota language and culture.

Similarly, the final group in the workshop focused on the potential for use of social media in language documentation or revitalization, specifically focusing on bots for language learning and analysis of social media text.

\subsection{Bots for Language Learning}

In discussing potential projects during the workshop, some members of the workshop who are actively learning their ancestral languages noted that they often had trouble coming up with words in the heritage language, and in maintaining their motivation to continue practicing the language. Thus, we created a prototype chatbot, ``Idiomata", that users can add to conversations that they would like to have in their language.  We implemented this for Seneca and Cayuga, two Haudenosaunee languages indigenous to what is now New York State and Ontario. Idiomata currently has two functionalities:

\begin{enumerate}\setlength{\itemsep}{-3pt}
\item It can be added to a chat between two speakers, and once it is added with the appropriate permissions, it will monitor the number of words the conversants use in each language. The conversants can ask ``@Idiomata my score", and the bot will inform the speakers of how many words they used in each language.
\item Second, the bot can propose translations. The speakers can write ``@Idiomata translate XXX", and a translation will be provided for the word XXX.
\end{enumerate}

\begin{figure}[t]
\centering
\includegraphics[width=\columnwidth]{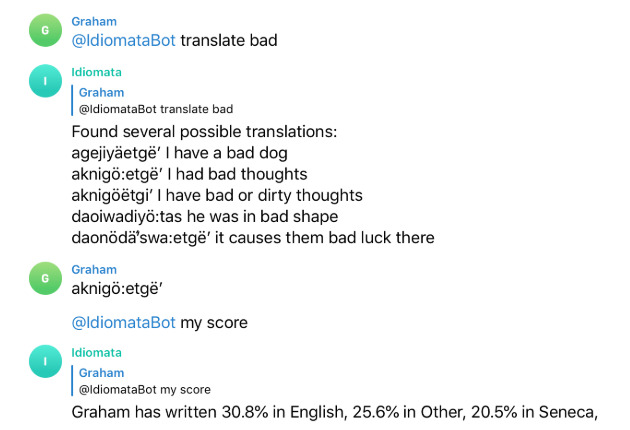} 
\vspace{-3mm}
\caption{Usage of the Idiomata bot to search for translations and obtain scores regarding number of words spoken.
\label{fig:idiomata1}}
\vspace{-3mm}
\end{figure}

An example of both functionalities is shown in the screenshot in Figure \ref{fig:idiomata1}.
In order to ensure that this functionality is easy to implement even for languages where there is a paucity of resources, both of the above functionalities are based only on a bilingual dictionary between the language the speakers would like to be conversing in, and a language of wider communication such as English. 

\begin{figure}[!ht]
\centering
\includegraphics[scale=0.51]{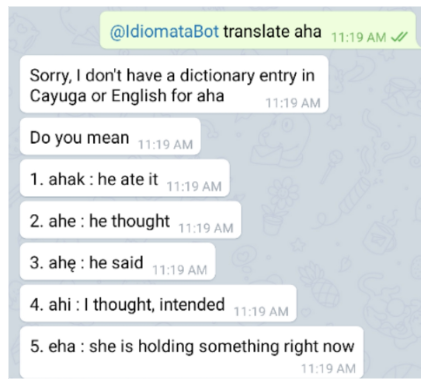} 
\vspace{-3mm}
\caption{An example of fuzzy dictionary matching.
\label{fig:idiomata2}}
\vspace{-3mm}
\end{figure}

Importantly for language learners, both functionalities have the capacity for fuzzy matching in case of spelling errors or morphological variation (Figure \ref{fig:idiomata2}). 
Specifically, the fuzzy matching is implemented based on edit distance algorithm for string matching. We evaluate the distance between the input of the user and the words in the dictionary of both the low-resourced language and English. If the user input is not in the dictionary, the chatbot will suggest five dictionary words with distances lower than a given threshold. 

\subsection{Analysis of Social Media Text}

A related endeavor of the workshop continued research into temporal codeswitching analyses in the Twitter domain. Focusing on Irish as a relatively well-resourced endangered language under somewhat of a current sociopolitical spotlight, but with the purpose of application to other endangered languages, work continued into more finely identifying features considered Irish versus belonging, lexically or by intent of the user, to some other language---for example, the ``Guarda" term semantically situating itself between the ``garda" Irish police service and the related English term ``guard." Such identifications are important to temporal tracking of endangered language decay (dropout) or shifts within popular multilingual social media platforms where certain languages may feature prominently and pressure shifts on behalf of enabled or sustained engagement. 

For these purposes, the continued and future work of the temporal language shift modeling forecasts integration of the aforementioned chatbot project on behalf of both tracking language shifts within individual conversation threads, not just at the level of tracking language shift over a user account lifespan, as well as on behalf of testing models of language shift within social media settings more so under the researchers' control, such as in a chatbot setting involving multiple users or a user and bot. 

\section{Conclusions}

During the workshop teams made significant progress on speech technology, dictionary management, teacher-in-the-loop technology, and social media. 
Participants responded that they found it quite productive and would like to continue in the following year. Based on this, the organizers are looking to hold the workshop again in the summer of 2020. While there were a variety of participants in the workshop, there was a slant towards technologists given the location of and organization of the workshop, suggesting that perhaps future workshops should be located in a place more conducive to encouraging participation from language community members, either in their territory, or co-located with a conference such as CoLang. Three workshop participants (Anastasopolous, Cruz, and Neubig) are also planning a tutorial on technology for endangered languages at the 2020 edition of the Conference on Computational Linguistics (COLING).

\section{Acknowledgments}
The authors thank all participants in the workshop. The workshop and this report were supported in part by NSF Award No.~1761548.


\section{Bibliographic References}
\bibliographystyle{lrec}
\bibliography{ltldr}

\section{Language Resource References}
\bibliographystylelanguageresource{lrec}
\bibliographylanguageresource{ltldr}

\end{document}